%% file: main.tex
\documentclass[a4paper,11pt]{article}

\usepackage[T1]{fontenc}
\usepackage[utf8]{inputenc} 
\usepackage[english]{babel}

\usepackage{amsmath, amssymb}
\usepackage{graphicx}
\usepackage{booktabs}
\usepackage{tabularx}
\usepackage{array}
\usepackage{float}
\usepackage{pifont}
\usepackage{makecell}
\usepackage{subcaption}
\usepackage{hyperref}
\usepackage[a4paper, margin=2.5cm]{geometry}
\usepackage{cite}

\title{\textbf{\LARGE Literature Review on Emerging Trends in Pseudo-Label Refinement for Weakly Supervised Semantic Segmentation with Image-Level Supervision}}

\author{
\begin{tabular}{c c}
    Zheyuan Zhang & Wang Zhang \\
    Informatics Institute & School of Artificial Intelligence \\
    University of Amsterdam & Beijing Normal University \\
    Amsterdam, Netherlands & Beijing, China \\
    \texttt{mickey.zhang@student.uva.nl} & \texttt{zhangwang@mail.bnu.edu.cn}
\end{tabular}
}

\begin{document}
\date{}
\maketitle

\begin{abstract}
Unlike fully supervised semantic segmentation, weakly supervised semantic segmentation (WSSS) relies on weaker forms of supervision to perform dense prediction tasks. Among the various types of weak supervision, WSSS with image-level annotations is considered both the most challenging and the most practical, attracting significant research attention. Therefore, in this review, we focus on WSSS with image-level annotations. Additionally, this review concentrates on mainstream research directions, deliberately omitting less influential branches.

Given the rapid development of new methods and the limitations of existing surveys in capturing recent trends, there is a pressing need for an updated and comprehensive review. Our goal is to fill this gap by synthesizing the latest advancements and state-of-the-art techniques in WSSS with image-level labels. 

Basically, we provide a comprehensive review of recent advancements in WSSS with image-level labels, categorizing existing methods based on the types and levels of additional supervision involved.  We also examine the challenges of applying advanced methods to domain-specific datasets in WSSS—a topic that remains underexplored. Finally, we discuss the current challenges, evaluate the limitations of existing approaches, and outline several promising directions for future research. This review is intended for researchers who are already familiar with the fundamental concepts of WSSS and are seeking to deepen their understanding of current advances and methodological innovations. 

\end{abstract}


\input{1-Section.tex}
\input{2-Section.tex}
\input{3-Section.tex}

\input{4-Section.tex}

\input{5-Section.tex}

\newpage
\bibliographystyle{abbrv}
\bibliography{sample.bib}

\end{document}

%% file: 1-Section.tex
\newpage
\section{Introduction}
While fully supervised learning methods have achieved impressive results, these methods require a large scale of training images with pixel-level annotations, which is expensive and time-consuming, making it hard to be used in real world scenarios. For domain specific tasks, accurate labeling for training a fully supervised segmentation model requires professional knowledge.
To alleviate the burden of labeling, weakly supervised semantic segmentation with only image-level labels becomes appealing as a practical alternative, where class activation map is used to generate pseudo labels.

Nevertheless, due to the limited nature of image-level supervision, CAMs often suffer from several issues. First, co-occurrence issues (e.g., a chimney appearing in a photo labeled as smoke) can lead to incorrect activation. Second, CAMs typically highlight only the most discriminative regions of an object, resulting in partial coverage of the target area.                                                                 To To address these limitations, various refinement techniques have been proposed.  These techniques can be applied both in CAM generation and after CAM generation. Here we only discuss the techniques that aims at optimizing CAM generation in the training process, as post-processing is sensitive to noise in the initial CAM and cannot handle fundamental issue like spurious correlations.

The mainstream methodology for pseudo labels refinement under image-level supervision is to introduce more supervision to guide the classifier to generate more precise and complete CAMs. We survey recent advanced works and  divide them as external supervision and internal supervision method. For each type of supervision, we demonstrate the underlying methodology and techniques that applied.

\begin{figure}[h!]
	\centering
	\includegraphics[width=0.6\linewidth]{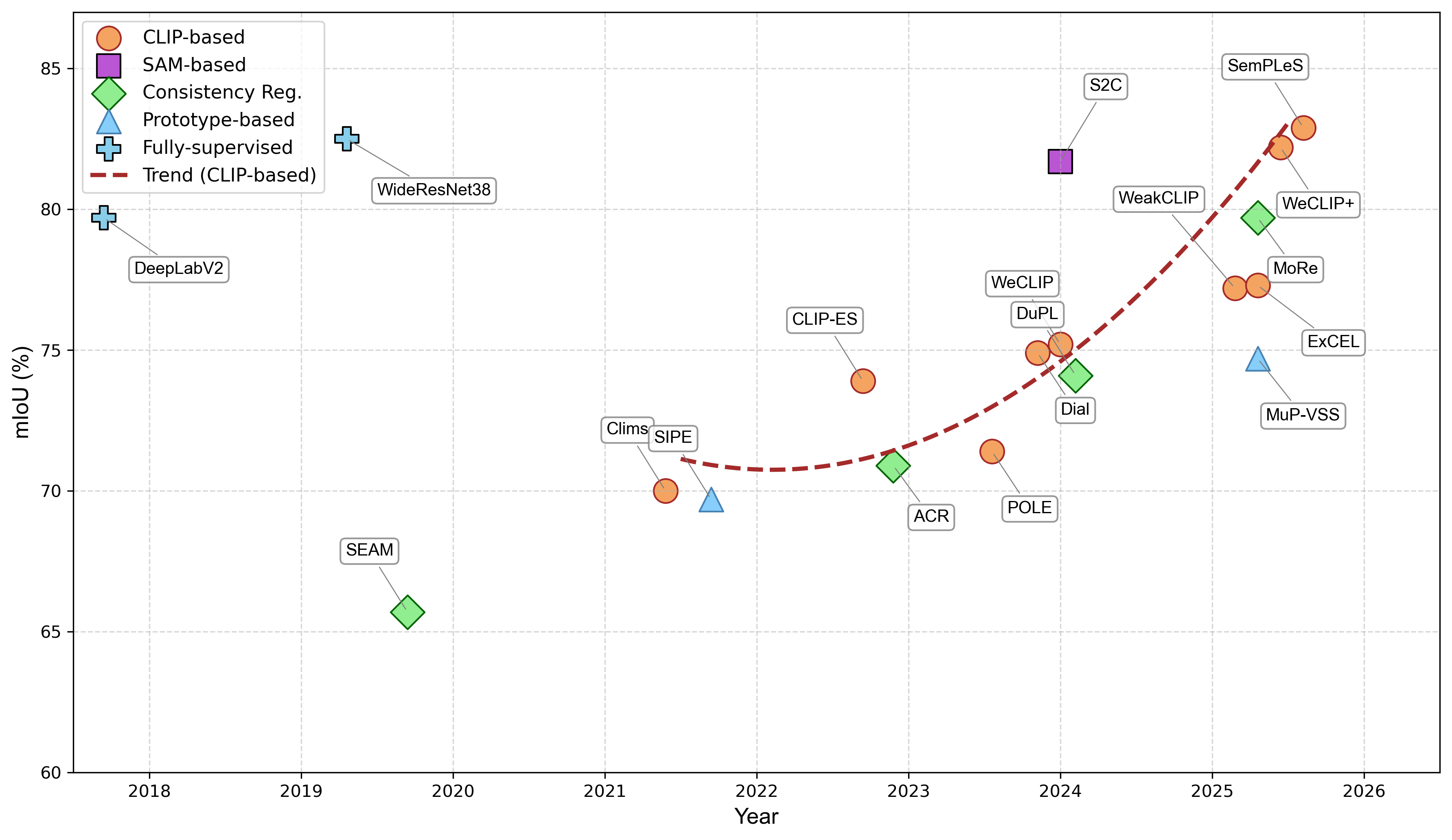}
	\caption{The performance of recent WSSS works on VOC test set.}
	\label{fig:previous methods}
\end{figure}

Our main contribution are summarized as:

\begin{itemize}
\item We review the most recent and advanced WSSS methods based on image-level annotations in depth and propose a new taxonomy to categorize them into internal and external supervision strategies.
\item We investigate the challenges of applying these methods to domain-specific datasets, which are often underrepresented in current benchmarks.
\item We identify several promising directions for future research in this field.
\end{itemize}

%% file: 2-Section.tex
\section{Related Work}
\label{chapter2}
\textit{}

The number of review papers specifically focused on WSSS is quite limited. Most existing surveys tend to cover fully supervised or semi-supervised methods alongside WSSS, rather than treating WSSS as an independent area of study. For example, \cite{zhang2020survey} discuss both semi and weakly supervised semantic segmentation. As for literature that focus on WSSS. Lu et al.~\cite{lu2018survey} discuss both MIL and CAM-based methods, but lack a clear categorization of the characteristics that distinguish different approaches. In contrast, our review focuses exclusively on CAM-based pseudo-label generation, as it remains the mainstream methodology in WSSS. Zhu et al.~\cite{zhu2023survey} categorize WSSS methods based on various types of weak annotations—such as image-level labels, point annotations, bounding boxes, and scribbles. However, we concentrate solely on image-level supervision, as it is the most cost-effective and practical annotation type, making it more suitable for real-world applications.

Most of the aforementioned surveys primarily explain the basic concepts and general procedures of WSSS, without delving into the critical details of pseudo-label optimization, which is a key challenge in the field. Additionally, many of these works are becoming outdated due to the rapid development of new techniques.
Therefore, this review fills an important gap. We also highlight a recent related work by Chen et al.~\cite{chen2025weakly}, which explores the potential of applying foundation models to WSSS. While their work focuses primarily on external supervision, we extend this discussion by exploring both external supervision (e.g., foundation models) and internal supervision (e.g., consistency regularization), offering a more holistic perspective on pseudo-label refinement.

%% file: 3-Section.tex
\section{Study Design}
\label{chapter3}

We primarily used Google Scholar and the ACM Digital Library to search for relevant literature. All retrieved references were organized and managed using Zotero. We identified a set of papers as our primary studies based on their relevance to the topic.
The Search String is : \textit{“Weakly supervised semantic segmentation” AND “image-level” AND “Class Activation Map”}

\textbf{Scope:}
The reviewed works span the period from 2020 to 2025. To ensure both quality and relevance, we applied the following inclusion criteria: for papers published between 2020 and 2023, a minimum of 10 citations was required. Papers published in 2024 or later were included regardless of citation count, recognizing that newer publications may not yet have accumulated citations.

\textbf{Snowballing Strategy:} 
To supplement our primary search, we employed a snowballing approach. For each selected primary study, we examined the references cited in their related work sections—particularly those used for performance comparisons—to identify additional relevant studies. Furthermore, we analyzed methodological similarities in result comparisons to uncover pipelines of the same type.

\textbf{Research Questions:}
\begin{itemize}
    \item \textbf{RQ1:} What are the emerging trends in pseudo-label refinement for weakly supervised semantic segmentation using image-level annotations?
    \item \textbf{RQ2:} What are the current challenges and underexplored areas in weakly supervised semantic segmentation with image-level annotations?
\end{itemize}

%% file: 4-Section.tex
\section{Emerging Trends}
Due to the insufficient supervision in WSSS, a natural methodology is to introduce additional supervision to regularize the model to learn accuracy and compact CAMs.
We categorize recent studies based on the type of additional supervision introduced—internal supervision and external supervision—both of which aim to bridge the supervision gap by leveraging auxiliary knowledge and regularization techniques.

\subsection{Internal supervision}
\label{chapter 4}
To addressing the aforementioned challenges, one methodology is to seek for additional internal supervision. The mainstream way is to use consistency regularization or prototype contrastive learning to achieve internal supervision without external supervision or knowledge. 

\subsubsection{Consistency Regularization}
The methodology of consistency regularization involves identifying existing implicit representation discrepancies or deliberately designing such discrepancies, and then enforcing consistency across them. This approach has been widely adopted in prior work. Based on the consistency level ,we can divide the existing work into several categories.

\textbf{Image-level Consistency:} This type of method is based on the hypothesis that models should maintain semantic consistency across different views(e.g.,data augmentation, geometric transformation \cite{misra2020self}) of an image. SEAM \cite{wang2020self} enforces CAMs learned from different augmented version of input images to maintain consistency. ACR \cite{sun2023all} is similar to SEAM but introduce affinity consistency to offer better supervision.

\textbf{Patch-level Consistency:}
CPN \cite{zhang2021complementary} find the discrepancy from activated patch and its complementary patch from the same image and narrow down the gap between them. CPN refines the quality of pseudo labels by addressing partial coverage issue, but this kind of consistency regularization cannot address spurious correlations issue. ToCo \cite{ru2023token} observes that there exists discrepancy between patch tokens and between global class token and local class token and design two token contrast modules to facilitate the representation consistency.

\textbf{Class-Patch-level Consistency:}
Unlike ToCo, which treats class tokens and patch tokens independently and enforces consistency separately, MoRe \cite{yang2025more} argues that it is essential to impose regularization between class and patch tokens. To achieve this, they model attention as a graph and apply regularization to the interactions at the class-patch level. This design leverages the complementary roles of the two token types: class tokens may introduce semantic bias, while patch tokens preserve fine-grained spatial details, making their joint modeling both intuitive and effective.


\textbf{Cross-model Consistency:}
This type of methods is to train two parallel model with different view or different initial parameter or manually impose discrepancy by devising a loss function and build discrepancy to generate diverse CAMs by co-training.
For example, in order to impose two classifier to have distinct CAMs, \cite{zhang2020splitting} proposes a discrepancy loss to regularize the two classifier. 
Usage \cite{peng2023usage} build discrepancy by adjust the network structure using dropout technique.
DuPL \cite{wu2024dupl} is a end to end method that leverages dual student network with same architecture and also use a discrepancy loss to constrain the two student model to generate diverse CAMs and then use the CAM from another student to guide the model to generate segmentation masks. 

\begin{table}[H]
\small
\centering
\caption{Comparison of methods consistency regularization (I: image label, S: saliency map).}
\label{tab:methods_comparison}
\begin{tabular}{p{3.7cm} p{1.9cm} p{1.2cm} 
>{\centering\arraybackslash}p{0.6cm} p{1.5cm} 
>{\centering\arraybackslash}p{1.8cm} p{2.2cm}}
\toprule
\textbf{Consistency Level} & \textbf{Method} & \textbf{Venue} & \textbf{Code} & \textbf{Backbone} & \textbf{Supervision} & \textbf{Dataset} \\
\midrule
Image-level  & SEAM \cite{wang2020self} & CVPR'20 & \checkmark & ResNet38  & I & VOC,COCO \\
 & ACR \cite{sun2023all} & ICCVW'24 & \checkmark & ViT-B  & I & VOC,COCO \\
 \midrule
Patch-level & CPN \cite{zhang2021complementary} & ICCV'21 & \checkmark & ResNet38 & I &  VOC \\
 & ToCo \cite{ru2023token} & CVPR'23 & \checkmark & ViT-B & I & VOC, COCO \\
 \midrule
Class-patch level & MoRe \cite{yang2025more} & AAAI'25 & \checkmark & ViT-B & I & VOC,COCO\\
\midrule
Cross-model &DuPL \cite{wu2024dupl} & CVPR'24 & \checkmark & ViT-B & I & VOC,COCO\\
 & S\&M\cite{zhang2020splitting} & ECCV'20 & \ding{56} & VGG16 & I+S & VOC \\
& Usage \cite{peng2023usage} & ICCV'23 & \checkmark & ResNet50 & I & VOC,COCO \\
 & BECO \cite{rong2023boundary} & CVPR'23 & \checkmark & ResNet50 & I & VOC, COCO \\

\bottomrule
\end{tabular}
\end{table}

\subsubsection{Prototypical Contrastive Learning}
The basic idea is to generate prototypes like class prototype that can represent features, or specific attributes of an object and then use contrastive learning, where prototypes are served as anchors,to reduce knowledge bias. Memory bank is commonly used for storing massive and diverse prototypes. This technique can also be regarded as consistency regularization. However, as prototype learning plays a important role here, we distinguish it from conventional consistency-based methods.
Another reason for this separation is the growing attention prototype learning has received in recent years—it is emerging as a new driving force in the WSSS community, offering a promising direction for improving pseudo-label quality through structured and discriminative representation learning.

Prototypical Contrastive Learning is first proposed in \cite{li2020prototypical} to address the weakness of instance-wise contrastive learning in few-shot learning.
Unlike few-shot learning, there are no pixel-level annotations, making prototype generation more challenging.
In this weaker setting, prototype can be generated by feature similarity or using a pretrained segmentation model \cite{kweon2024sam}. 

Several studies, such as \cite{chen2023extracting}, demonstrate that prototype learning alone can achieve strong performance in weakly supervised semantic segmentation. Specifically, \cite{chen2023extracting} mitigates the weakness of classifier CAMs with prototype based CAMs. Building upon this foundation, contrastive learning has been introduced to further enhance feature representations. For instance, Du et al. \cite{du2022weakly} employs prototypical contrastive learning to enforce consistency between pixel-level features and their corresponding class prototypes, leading to more discriminative and robust representations.
\cite{das2023weakly} adopts prototype learning in domain adaptation and aligns the pixel level feature with prototypes. However, pixel-level relation modeling may put over-regularization to the model and resulting in sub-optimal performance. SIPE\cite{chen2022self} in contrast leverages image-specific prototype to generate more complete CAMs. 
Despite its advantages, the  class prototype derived from classification loss still suffer from partial coverage because of  the inherent supervision gap. Thus the aforementioned methods still suffer from incomplete CAMs.
To address this issue, PSDPM \cite{zhao2024psdpm} proposes a method to estimate potential foreground pixels, which helps regularize the model by encouraging attention to less discriminative foreground regions during the learning of discriminative features.
Considering that pixels belonging to the same class can exhibit significant variation even within the same image, prototypes that incorporate contextual knowledge have emerged as a more robust representation strategy. RCA \cite{zhou2022regional} gathers diverse relational contexts to enrich prototype semantic representations. CPAL \cite{tang2024hunting} proposes a strategy to select candidate contextual prototypes as neighbors for the current instance prototype, aiming to mitigate the bias introduced by relying on a single prototype when incorporating contextual knowledge. \cite{duan2025multi} utilize patch tokens to learn global class-specific correlation to avoid the use of biased class token.

Given this limitation, an important research direction in prototype learning is to explore how to effectively model and leverage the relationship between class tokens and patch tokens. Specifically, learning how class-level semantics interact with localized visual patterns can lead to more complete and accurate prototypes that better align with pixel-level segmentation objectives.  In addition, \cite{wang2025pot} explores the use of optimal transport for prototype selection. Several other works also leverage foundation models such as SAM or CLIP to construct textual prototypes \cite{yang2024foundation} or visual prototypes \cite{tang2025prototype}, opening new directions for prototype learning in WSSS.

\begin{table}[htbp]
\small
\centering
\caption{Comparison of prototype-based method. (I: image label, T: CLIP, S: SAM)}
\label{tab:clip}
\begin{tabular}{p{4cm} p{3.1cm} p{1.7cm} 
>{\centering\arraybackslash}p{0.8cm} 
>{\centering\arraybackslash}p{2cm} p{2.5cm}}
\toprule
\textbf{Type} & \textbf{Method} & \textbf{Venue} & \textbf{Code}& \textbf{Supervision} & \textbf{Dataset} \\
\midrule

Pixel-to-Prototype & Du et al. \cite{du2022weakly}&CVPR'22&\checkmark &I&VOC,COCO\\
 \midrule
Context-aware Prototype & CPAL \cite{tang2024hunting}&CVPR'24&\checkmark &I&VOC,COCO\\
 & RCA \cite{zhou2022regional}&CVPR'22&\checkmark &I&VOC,COCO\\

 \midrule
 Modality-bridged 
 & PBIP \cite{tang2025prototype} & CVPR'25 & \checkmark & I+T & Medical imaging\\
& FMA \cite{yang2024foundation} & WACV'24  & \checkmark  &  I+T & VOC,COCO  \\
\bottomrule
\end{tabular}
\end{table}

\subsection{External supervision}
This line of research introduces natural language supervision or leverages the zero-shot capabilities of pre-trained foundation models to refine the quality of pseudo labels. These approaches can be broadly categorized into CLIP-based and SAM-based methods. While there are also emerging LLM-based approaches (e.g., \cite{wu2025prompt}), they are still in the early stages and have not yet received significant emphasis in the field. Therefore, we do not explore them in this work.

\subsubsection{CLIP-based}
Contrastive Language-Image Pre-training (CLIP) \cite{radford2021learning}  has demonstrated a strong alignment between visual concepts and textual descriptions, showcasing powerful cross-modal representation capabilities.

Recent works~\cite{zhang2024frozen,xie2022clims,lin2023clip,lin2024semples,xu2023learning} introduce CLIP primarily to design text-driven background suppression modules  through CAMs generated by CLIP to enhance the quality of pseudo labels . This is crucial since relying solely on image-level supervision often leads classifiers to learn biased knowledge. 
Another line of work is to use CLIP directly as a backbone network~\cite{zhang2024frozen}. 

The typical procedure to generate CAMs using CLIP includes the following steps:
	1.	Input the image into the image encoder to extract image features.
	2.	Generate textual prompts for each class label and feed them into the text encoder to obtain corresponding text features.
	3.	Compute the similarity (e.g., cosine similarity) between the image and text features. These similarities are then used as classification weights to generate CAMs.
    
\textbf{Manual Prompting:}
CLIMS~\cite{xie2022clims} enhances CLIP by maximizing similarity between image and text features and further mitigates co-occurrence issues by jointly optimizing for both target classes and their commonly co-occurring backgrounds. CLIP-ES~\cite{lin2023clip} also utilizes CLIP features, but both CLIMS and CLIP-ES rely on manually defined prompts (e.g., “a photo of {}”), which are overly simplistic and not optimal for adapting CLIP to downstream tasks. Moreover, manual prompt engineering requires domain expertise and does not fully eliminate human effort. 

\textbf{Prompt Learning:}
CAM generation has been found to be highly sensitive to the choice of text prompts~\cite{murugesan2024prompting}. To better utilize CLIP, one promising direction is to apply prompt learning to automatically optimize text inputs without human intervention. For instance, SemPLeS~\cite{lin2024semples} employs contrastive prompt learning to construct a sequence of learnable background prompts that guide the mask generator to exclude semantically irrelevant regions. This approach effectively addresses co-occurring background issues and strengthens image-text alignment. 

In domain-specific tasks, \cite{wang2024creative} fine-tunes CLIP on curated pairs of remote sensing images, while  \cite{zheng2024exploring} proposes weakly supervised prompt learning for generating medical-specific prompts, enabling CLIP to learn more transferable representations. These works suggests that directly adopt vanilla CLIP in domain-specific tasks may yields limited effectiveness.

\begin{table}[htbp]
\small
\centering
\caption{Comparison of CLIP-based method. (I: image label, T: CLIP, S: SAM)}
\label{tab:clip}
\begin{tabular}{p{3.6cm} p{3.1cm} p{1.2cm} 
>{\centering\arraybackslash}p{0.8cm} p{1.4cm} 
>{\centering\arraybackslash}p{1.7cm} p{2.5cm}}
\toprule
\textbf{Type} & \textbf{Method} & \textbf{Venue} & \textbf{Code} & \textbf{Backbone} & \textbf{Supervision} & \textbf{Dataset} \\
\midrule
Manual prompt & CLIMS~\cite{xie2022clims} & CVPR'22 & \checkmark & ResNet50& I+T & VOC, COCO  \\

 & CLIP-ES \cite{lin2023clip} & CVPR'23 & \checkmark & ResNet101 & I+T & VOC, COCO \\
 \midrule

Prompt learning & POLE \cite{murugesan2024prompting} & WACV'24 & \checkmark & ResNet50 & I+T & VOC,COCO \\
 & SemPLeS \cite{lin2024semples} & WACV'25 & \checkmark & Swin-L & I+T & VOC, COCO \\ 
& MedPrompt \cite{zheng2024exploring}& PR'24 & \checkmark & ViT & I+T & Medical imaging \\
& FMA \cite{yang2024foundation} & WACV'24 & \checkmark & Swin-L& I+T+S & VOC, COCO \\
 & MMCST \cite{xu2023learning} & CVPR'23 & \ding{56} & ResNet38 & I+T & VOC, COCO \\
\midrule
CLIP as backbone & WeCLIP \cite{zhang2024frozen} & CVPR'24 & \checkmark & CLIP & I+T & VOC,COCO \\

  \midrule
Domain-specific & RS-TextWS-Seg \cite{wang2024creative} & TGRS'24 & \ding{56} & ResNet38 & I+T+S & Reomte sensing \\
& CARB \cite{kim2024weakly} & AAAI'24 & \checkmark & CLIP & I+T & Cityscapes \\
\bottomrule
\end{tabular}
\end{table}

\subsubsection{SAM-based}
Thanks to SAM's powerful zero-shot ability, masks generated by SAM shows accurate boundary and complete spatial consistency. Thus SAM is mainly adapted during inference time through post-processing \cite{chen2023segment}.In this approach, each mask generated by SAM is assigned the class label corresponding to the pseudo-label with the highest overlap. Other strategies have also been proposed to assign class labels to class-agnostic masks. For example, class labels can be expressed in natural language and encoded using CLIP. The resulting embeddings are then used to guide SAM, where the predicted masks for each class serve as bounding box prompts to generate class-specific masks.
However, these post-processing methods are sensitive to noise and may amplify errors instead of correcting them. Meanwhile, SAM often segments objects composed of multiple components, such as bicycles, into separate parts, rather than capturing them as a single, unified instance.

Recently S2C \cite{kweon2024sam} has explored the direct transfer of knowledge of SAM to the classifier using prototypical contrastive learning to address these issues. However, while SAM already provides strong performance, research attention has increasingly shifted toward cross-modality methods like CLIP-based approaches, due to the substantial performance gap that still exists between CLIP-based, SAM-based and fully supervised methods.

\begin{table}[htbp]
\small
\centering
\caption{Comparison of SAM-based method. (I: image label, T: CLIP, S: SAM)}
\label{tab:clip}
\begin{tabular}{p{3.6cm} p{3.1cm} p{1.7cm} 
>{\centering\arraybackslash}p{0.8cm} 
>{\centering\arraybackslash}p{2cm} p{2.4cm}}
\toprule
\textbf{Type} & \textbf{Method} & \textbf{Venue} & \textbf{Code}& \textbf{Supervision} & \textbf{Dataset} \\
\midrule
Post-processing & SEPL \cite{chen2023segment} & NeurIPSW'23 & \ding{56} &  I+S & VOC,COCO  \\
 \midrule

Knowledge Transfer & S2C \cite{kweon2024sam} & CVPR'24 & \checkmark & I+S & VOC,COCO \\
\bottomrule
\end{tabular}
\end{table}

\subsection{Discussion}
Through the investigation of recent WSSS works, we have some insights:

(1)How to enforce sufficient diversity to their representations is the key to consistency regularization. While consistency regularization has shown promise, most of the previous works mainly focus on discrepancy in a single model with different view or cross-model consistency. However, very few works explore cross-architecture consistency. And the existing works lacks an effective way to exploit this discrepancy. For example, \cite{park2024precision} simply fuses the cams generated from CNN and ViT using AND and OR operation. This may due to the fact that  heterogeneous feature from different architecture model is hard to be leveraged effectively.

(2)  Meanwhile, SAM can be adopted in domain specific task with much efforts while CLIP may failed in such case. As shown in Table~\ref{tab:clip}, leveraging natural language supervision is effective on general datasets such as PASCAL VOC and COCO. However, the language-based models may struggle to be generalized. For domain-specific applications (e.g., remote sensing or medical image), fine-tuning CLIP proves more effective than relying on its zero-shot capabilities, as vanilla CLIP lacks the necessary aligned image-text embeddings for specialized contexts. 

(3) As for domain specific task, internal supervision like prototype learning seems to be a very powerful way to extract rich semantic representation of domain specific tasks.

%% file: 5-Section.tex
\section{Summary and Future direction}
In conclusion, we propose a new taxonomy for WSSS and analyze the methodologies within each category, highlighting their respective advantages and limitations. We also identify several promising directions for future research.

\noindent\textbf{Future Directions}
The future directions are as follows:
\begin{enumerate}
    \item For consistency regularization, effectively aligning heterogeneous features across different model architectures remains a major challenge. The core difficulty lies in finding an effective way to aggregate diverse and incompatible knowledge from heterogeneous models.
    
    \item The use of foundation models in WSSS is still in its preliminary stages, leaving ample room for further exploration. For domain-specific tasks, adapting foundation models to WSSS remains largely unexplored. The main challenge lies in how to effectively leverage transferable knowledge (e.g., from SAM) or apply prompt learning techniques (e.g., for CLIP) to align domain-specific textual cues with visual representations. Meanwhile, for CLIP-based method, there is also a need for weakly supervised prompt learning to automatically generate domain-specific prompts.
    
    \item From our perspective, for domain-specific task, internal supervision like prototype contrastive learning appears to be a highly promising direction for advancing WSSS. A key research challenge lies in constructing representative and robust prototypes that capture intra-class variations while remaining semantically consistent. For example, leverage SAM to generate unbiased prototype is worth trying. Additionally, modality-bridged prototypical contrastive learning, which learns modality-invariant prototypes serving as semantic anchors to align different modalities, offers further potential to enhance cross-modal consistency in WSSS.
\end{enumerate}